\documentclass[runningheads]{llncs}

\usepackage{graphicx}
\usepackage{amsmath}
\usepackage{subcaption}
\usepackage[ruled]{algorithm2e}
\usepackage{amssymb}

\usepackage{ctable}
\usepackage{xcolor}
\begin{document}
%===========================================================

\title{Introspective Learning by Distilling Knowledge from Online Self-explanation} % Replace your paper's title here
\titlerunning{Introspective Learning} % Replace an abstracted version of your paper's title here

%===========================================================

\author{Jindong Gu$^{1,2}$ \and  Zhiliang Wu$^{1,2}$ \and  Volker Tresp$^{1,2}$}

%
%Please include author names in full in the paper, 
%If any authors have names that can be parsed into FirstName LastName in multiple ways, please include the correct parsing, in a comment to the volume editors:
%\index{Lastnames, Firstnames}

\authorrunning{Jindong Gu et al.} % A shorter version of authors' name

%===========================================================

\institute{University of Munich, Munich, Germany \\ \and
Siemens AG, Corporate Technology, Munich, Germany \\ }

\maketitle

%===========================================================
\begin{abstract}
In recent years, many explanation methods have been proposed to explain individual classifications of deep neural networks. However, how to leverage the created explanations to improve the learning process has been less explored. As the privileged information, the explanations of a model can be used to guide the learning process of the model itself. In the community, another intensively investigated privileged information used to guide the training of a model is the knowledge from a powerful teacher model. The goal of this work is to leverage the self-explanation to improve the learning process by borrowing ideas from knowledge distillation. We start by investigating the effective components of the knowledge transferred from the teacher network to the student network. Our investigation reveals that both the responses in non-ground-truth classes and class-similarity information in teacher's outputs contribute to the success of the knowledge distillation. Motivated by the conclusion, we propose an implementation of introspective learning by distilling knowledge from online self-explanations. The models trained with the introspective learning procedure outperform the ones trained with the standard learning procedure, as well as the ones trained with different regularization methods. When compared to the models learned from peer networks or teacher networks, our models also show competitive performance and requires neither peers nor teachers.
\end{abstract}

\section{Introduction}
When human subjects imagine visual objects without the actual sensory stimulus, there is still activity in their visual cortex \cite{Kastner1999IncreasedAI}. The evidence implies that the internal representations in our visual cortex are used to reason about images. Similarly, when explaining classifications of neural networks, the existing model-aware explanation methods create explanations based on internal feature representations of neural networks. The explanation of each prediction identifies the relationship between input features and outputs. For instance, in image classification, saliency maps are proposed to explain how each input pixel is relevant to different classes,  \cite{Simonyan2013DeepIC,springenberg2014striving,shrikumar2017learning,selvaraju2017grad,gu2019contextual} to name a few.

It has been shown that human decision-making performance is related to the explanations during introspection in the brain \cite{Wilson1991ThinkingTM,Leisti2016TheEO}. In the community, there is no clear definition of introspective learning. In this work, we define \textit{introspective learning} as learning with explanations created from the underlying model itself. In recent years, the created explanations are mainly used to gain trust from users in real-world applications and help machine learning experts understand the predictions. How to leverage the explanations to improve the model itself has not been well explored.

Broadly speaking, learning with explanations falls into the framework of learning using privileged information \cite{Vapnik2009ANL,Vapnik2015LearningUP}. In classical supervised learning, the training data are $\{(\pmb{x}_1, \pmb{y}_1), \cdots , (\pmb{x}_n, \pmb{y}_n)\}$ where $(\pmb{x}_i, \pmb{y}_i)$ is an input-label pair. In the framework of learning using privileged information, additional information $\pmb{x}^*_i$ about $(\pmb{x}_i, \pmb{y}_i)$ is provided by an intelligent teacher. The training data are therefore formed by a collection of triplets
\begin{equation}
\{(\pmb{x}_1, \pmb{x}^*_1, \pmb{y}_1), \cdots , (\pmb{x}_n, \pmb{x}^*_n, \pmb{y}_n)\}
\end{equation}
For instance, $\pmb{x}^*_i$ can be the softened outputs of a teacher model \cite{LopezPaz2015UnifyingDA}. 
In this work, we consider the explanation $\pmb{E}_i$ created for the classification of $(\pmb{x}_i, \pmb{y}_i)$ as the privileged information.

When the softened outputs of a powerful teacher model are provided as the privileged information, the learning process is known as knowledge distillation (KD) from Hinton et al. \cite{hinton2015distilling}, where a student network is trained to match its softened outputs to teacher's. 

In this work, we first investigate the effective components in the knowledge distillation. Our investigation shows that both the responses in non-ground-truth classes and class-similarity information in teacher's outputs contribute to the effectiveness of knowledge distillation. Motivated by the conclusions, we propose a novel training procedure to train networks by leveraging online self-explanations. 

For the classification of the $i$-th sample $\pmb{x}_i$, we can create one saliency map for each output class, i.e., $\pmb{E}_i = (\pmb{e}_i^1, \pmb{e}_i^2, \cdots, \pmb{e}_i^c)$ where $c$ is the total number of output classes. The explanation $\pmb{e}^j_i$ is a function of the input $\pmb{x}_i$, the underlying model $M$ and the target class $y^j_i$, namely $\pmb{e}^j_i = f_e(\pmb{x}_i, M, y^j_i)$. They contain neither additional human prior knowledge nor knowledge of a powerful teacher model. 

It is perhaps not obvious why explanations can be used to improve the model at all. Our hypothesis is that the explanations can be leveraged to help the training process to find a better local minimum in a way similar to KD. \cite{Ba2013DoDN} observes that small networks often have the same representation capacity as large networks. Compared with large networks, they are simply harder to train and find good local minima. We leverage explanations to help the training process to find a better local minimum.

The contribution of this work is two-fold. Firstly, we investigate effective components of KD. Secondly, we propose a way to implement introspective learning by distilling knowledge from self-explanations. The rest of the paper is structured as follows: Sec. \ref{sec:relate} reviews closely related work. Sec. \ref{sec:dark} and \ref{sec:learn} describe the two contributions respectively. Sec. \ref{sec:exp} conducts experiments to show that our proposed learning procedure can achieve competitive performance wihtout any teacher. Moreover, the last section concludes the paper.

\section{Related Work}
\label{sec:relate}
\textbf{Explanation:} The explanation of a classification for the $j$-th class $\pmb{e}^j_i$ illustrates how relevant the input features are to the prediction of the $j$-th class. Vanilla Gradient (\textit{Grad}) takes the gradient of an output neuron with respect to the input as an explanation \cite{Simonyan2013DeepIC}. Guided Backpropagation (\textit{GuidedBP}) differs from the Grad approach in handling ReLU layers, where only positive gradients are propagated through the ReLU layers \cite{springenberg2014striving}. \cite{shrikumar2017learning} multiplies the Grad with the input (\textit{Grad*Input}) to deal with numerical stability. \cite{Ancona2017TowardsBU} shows that \textit{Grad*Input} is equivalent to LRP \cite{bach2015pixel} and DeepLIFT \cite{shrikumar2017learning}. Another commonly used approach (\textit{Grad-CAM}) combines the feature maps linearly using the averaged gradients of each feature map \cite{selvaraju2017grad}.

All the approaches introduced above require only a single backpropagation to create an explanation, while SmoothGrad \cite{smilkov2017smoothgrad} and IntergratedGrad \cite{sundararajan2017axiomatic} require dozens of passes. Some other appraoches require two backpropagations, e.g., Contrastive LRP, although they create visually pleasant and class-discriminative explanations \cite{Gu2018UnderstandingID,gu2019saliency}. The explanations created by these saliency methods are often applied to diagnose models.

\textbf{Regularization:} We briefly introduce closely related regularization techniques. Label Smoothing \cite{Szegedy2014GoingDW} replaces the one-hot labels with soft labels where $\alpha$ is specified for the ground-truth class, and $(1-a)$ is evenly distributed to other classes ($\alpha$ is often set as 0.9). \cite{pereyra2017regularizing} proposed to penalize low entropy predictions, called Confidence Penalty. Another commonly used regularization is Dropout \cite{srivastava2014dropout}, which drops each neuron of a layer with a certain probability.

\textbf{Knowledge Distillation:} The knowledge of a teacher model can be used to guide the training of a student model. Many works focus on different forms of the dark knowledge extracted from the teacher, e.g., soft outputs \cite{bucilua2006model,hinton2015distilling}, Hints (intermediate representations) \cite{Romero2014FitNetsHF}, Activation-based. Gradient-based attention \cite{Zagoruyko2016PayingMA}, and Flow between layers \cite{yim2017gift}. 

Recent research shows that a more powerful teacher and a peer network can help train a student. \cite{Yuan_2020_CVPR} shows a poorly-trained teacher with much lower accuracy than the student can still improve the latter significantly, and a student can also enhance the teacher by reversing the KD procedure. \cite{Furlanello2018BornAN} distills knowledge from a peer network with a similar capacity. The peer network can even be with the same architecture, known as Born-Again Network. The work \cite{gu2020search} searches for better student architectures to learn distilled knowledge. Deep Mutual Learning \cite{zhang2018deep} train multiple students with the same architecture and regularize their outputs to be similar, where each student can outperform the one trained alone without mutual learning. In this work, we propose to learn from a student itself by distilling knowledge from its explanations.

\textbf{Effectiveness of KD:} \cite{hinton2015distilling} argues that the success of KD can be attributed to the information in the logit distribution that describes the similarity between output categories. \cite{Furlanello2018BornAN} investigates gradients of the loss in KD and shows that weighting training examples with the teacher's confidence can also improve the student model's performance. In this work, we conduct further experiments to reveal effective components in KD.

\section{Understanding Dark Knowledge}
\label{sec:dark}
Before proposing a better alternative to the knowledge from a teacher model, we investigate the dark knowledge that improves the student in KD.

Given a training example $(\pmb{x}_i, \pmb{y}_i)$ and the number of output classes $c$, the logits and the output probabilities of a network are $\pmb{a}_i = (a^1_i, a^2_i, \cdots, a^c_i)$ and $\pmb{p}_i = \text{softmax}(\pmb{a}_i) =(p^1_i, p^2_i, \cdots, p^c_i)$ respectively. The target used to compute the loss of this example is $\pmb{t}_i$. The target is the one-hot labels $\pmb{y}_i$ in supervised learning, while it is the output probabilities of a teacher model $\pmb{q}_i$ in KD.

In both cases, given a mini-batch $\{(\pmb{x}_i, \pmb{y}_i)\}_{i=1}^{b}$, the gradient of the cross-entropy loss $\mathcal{L} = \frac{1}{b} \sum_{i=1}^{b} ce(\pmb{p}_i, \pmb{t}_i)$ with respect to the logit $a^j$ is
\begin{equation}
\small
\frac{\partial \mathcal{L}}{\partial a^j} = \frac{1}{b} \sum_{i=1}^{b} \frac{\partial \mathcal{L}}{\partial a_i^j} = \frac{1}{b} \sum_{i=1}^{b} (p_i^j - t_{i}^j).
\end{equation}
Without loss of generality, the samples in the mini-batch can be divided into two parts: $\{(\pmb{x}_i)\}_{i=1}^{s}$ with the $j$-th class as the ground-truth class ($y^j_i=1$) and $\{(\pmb{x}_i)\}_{i=s+1}^{b}$ with other classes as ground-truth classes ($y^j_i = 0$). In the case $\pmb{t} = \pmb{y}$ (denoted as \textbf{CE} for cross entropy), the gradient can be formulated as
\begin{equation}
\small
\frac{\partial \mathcal{L}}{\partial a^j} = \frac{1}{b} (\sum_{i=1}^{s} (p_i^j - 1) + \sum_{i=s+1}^{b} p_i^j ).
\end{equation}

In the case of training with KD, $\pmb{t} = \pmb{q}$ (denoted as \textbf{KD}\footnote{Similar to \cite{Furlanello2018BornAN}, the temperature of softmax in KD is hidden to simplify notations.}),  the gradient can be formulated as
\begin{equation}
\small
\frac{\partial \mathcal{L}}{\partial a^j} = \frac{1}{b} (\sum_{i=1}^{s} (p_i^j - q_i^j) + \sum_{i=s+1}^{b} (p_i^j - q_i^j) ).
\label{equ:kd}
\end{equation}

\cite{Furlanello2018BornAN} argues that the teacher's confidence is the effective component of KD, and KD simply performs importance weighting. With one-hot labels, they propose to weigh each sample by the teacher's confidence in its maximum value,  i.e., Confidence Weighted by Teacher Max (\textbf{CWTM}).
\begin{equation}
\small
\frac{\partial \mathcal{L}}{\partial a^j} = \sum_{i=1}^{s} \frac{q_i^{max}}{\sum_{j=1}^{b} q_j^{max}}  * (p_i^j - 1) + \sum_{i=s+1}^{b} \frac{q_i^{max}}{\sum_{j=1}^{b} q_j^{max}} * p_i^j
\label{equ:cwtm}
\end{equation}
where $q_i^{max}$ is the maximum of teacher's outputs $\pmb{q}_i$. 

To check the effectiveness of class-similarity information, they randomly permute the teacher's non-argmax outputs of predicted distribution on a sample, i.e., Dark Knowledge with Permuted Predictions (\textbf{DKPP}). The $q_i^j$ in the second term of Equation \ref{equ:kd} is replaced by $q_i^k$, which is one of the permuted non-argmax outputs. By doing this, the information about the similarity between classes is removed.

We further design another two experiments to verify the effectiveness of teacher's confidence: 1) \textbf{CWTM-Permut}: the teacher's confidence (argmax outputs) is randomly permuted inside a batch, namely the weight of the sample $(\pmb{x}_i, \pmb{y}_i)$ in Equation \ref{equ:cwtm} is set as $ \frac{q_{k}^{max}}{\sum_{j=1}^{n} q_{j}^{max}}$ where $k$ is one of the permuted sample indexes; 2) \textbf{CWTM-Random}: $q_{i}^{max}$ in Equation \ref{equ:cwtm} is replaced by a random value selected from [$\beta$, 1) as the teacher's confidence.

\begin{table*}[t]
\small
        \centering
          \caption{Test Accuracy (\%) of CNNs on CIFAR10 dataset: the teacher CNN-10 achieves 90,04\% test accuracy, and the performance of students are listed. CWTM-P coresponds to CWTM-Permut, and CWTM-R to CWTM-Random.}
        \begin{tabular}{l c c c c c  c  c}
         \specialrule{.1em}{.05em}{.1em}
        &Model & CE & KD & CWTM  & CWTM-P & CWTM-R &  DKPP\\
          \hline
         Student1 & CNN-8 &  86,86{\tiny($\pm$0.30)}  &   88,48{\tiny($\pm$0.33)}  &  87,03{\tiny($\pm$0.26)} &  87,41{\tiny($\pm$0.32)}  & 87,22{\tiny($\pm$0.17)} &	87,73{\tiny($\pm$0.37)}  \\
          \hline
         Student2 & CNN-6   & 85,18{\tiny($\pm$0.32)} &  	86,55{\tiny($\pm$0.32)} &	85,65{\tiny($\pm$0.38)} &  85,74{\tiny($\pm$0.33)}  & 85,25{\tiny($\pm$0.31)} & 86,30{\tiny($\pm$0.42)}  \\
          \specialrule{.1em}{.05em}{.1em}
  \end{tabular}   
\label{tab:under_kd}
\end{table*}

We conduct experiments on CIFAR10 \cite{krizhevsky2009learning} using convolutional neural networks (CNN). The used CNNs consist of convolutional layers followed by max-pooling and ends with a fully connected layer. The teacher we use has ten convolutional layers, called CNN-10, and the two students are CNN-8 and CNN-6, respectively. The standard data augmentation is applied to the training data: 4 pixels are padded on each side, and a 32$\times$32 patch is randomly cropped from the padded images or their horizontal flip. The $\beta$ in CWTM-Random is 0.5, which ensures that $q_{i}^{max}$ corresponds to the maximal value. All models are trained with a batch size of 128 for 160 epochs using SGD with a learning rate of 0.01 and moment 0.9. The test performance is shown in Table \ref{tab:under_kd}. All the scores reported in this paper are averaged on five trials in forms of $\text{mean}(\pm \text{std})$.

In Table \ref{tab:under_kd}, KD outperforms the classical supervised learning CE, which indicates that the targets specified by the teacher do help the training of the students. CWTM outperforms the baseline CE. \textbf{The improvement cannot be attributed to the teacher's confidence since both CWTM-Permut and CWTM-Random also outperform CE}. Therefore, the teacher's confidence (i.e., the maximal output) is not one of the effective components of KD.

Furthermore, as shown in \cite{Furlanello2018BornAN}, CWTM does not always outperform CE. We also found that CWTM-Random is very sensitive to the choice of $\beta$. The non-uniform sampled importance leads to the possible improvement of CWTM variants. We conjecture that it might improve the model by helping the optimization to escape saddle points. Further exploration is left in future work.

DKPP removes the class-similarity information contained in teacher's outputs by permuting logits of non-ground-truth classes. \textbf{DKPP still clearly outperforms CE, which means the responses in non-ground-truth classes contribute to the effectiveness of KD.} Similar to Labels Smoothing, the responses in non-ground-truth classes prevent the student from becoming over-confident. Furthermore, we can also observe that there is a gap between DKPP and KD. Compared to KD, \textbf{the student performance does decrease by removing the similarity information in the teacher's outputs. The class-similarity information also contributes to the effectiveness of KD.} These two effective components are further discussed in Section \ref{sec:xai}.

The observations above are consistent with \cite{Furlanello2018BornAN}, but we gain more insights into the KD technique with more experiments. From these observations, we conclude that both non-ground-truth classes' responses and class-similarity information in the teacher's outputs contribute to the success of KD.

\section{Introspective Learning with Online Self-explanations}
\label{sec:learn}
Given a training sample $(\pmb{x}_i, \pmb{y}_i)$ and a neural network $f(\cdot)$ to be trained, the outputs of the forward inference are $\pmb{p}_i$. A saliency method is applied on the neural network to generate one explanation for each output class $\pmb{E}_i = (\pmb{e}_i^1, \pmb{e}_i^2, \cdots, \pmb{e}_i^c)$. 

One desired property of the explanations $\pmb{E}_i$ is that the explanations are class-discriminative. In terms of a single sample, the similarity between explanations corresponds to the similarity between output classes. For instance, if the similarity between the explanations of two output classes $sim(\pmb{e}_i^1, \pmb{e}_i^2)$ is high, the $1$st output class and the $2$nd one are similar to each other. The similarity can be measured with different metrics, such as cosine distance, multi-scale mean squared error \cite{wang2003multiscale}, and Wasserstein distance \cite{villani2009wasserstein}. In our experiments, we find that the simplest one (i.e., Cosine distance) is good enough to capture the similarity between two created explanations.

This section proposes an algorithm to train neural networks with the online self-explanations created during training.  As shown in Algorithm \ref{alg}, our LE (\textbf{L}earning with \textbf{E}xplanations) consists of two training stages.

\begin{algorithm}[h]
\SetAlgoLined
\KwData{training samples $\{(\pmb{x}_i, \pmb{y}_i)\}_{i=1}^{n}$, a smooth factor $\alpha$}
\KwResult{a well-trained neural network}
\vspace{-0.1cm}
\hrulefill \\
 Stage 1 (Warm-up Training):  \\
\hspace{0.55cm}  train the network with one-hot labels $\mathcal{L}(\pmb{p}_i, \pmb{y}_i)$\;
\vspace{0.1cm}

 Stage 2 (Training with Online Self-explanations): \\
\begin{tabular}{c  p{8cm}}
 & \For{each epoch}{
          make forward inference $\pmb{p}_i = f(\pmb{x}_i)$\;
          generate explanations $\pmb{E}_i = (\pmb{e}_i^1, \pmb{e}_i^2, \cdots, \pmb{e}_i^c)$\;
          generate targets $\pmb{q}_i = (q_i^1, q_i^2, \cdots, q_i^c)$  with $\pmb{E}_i$ and $\pmb{y}_i$\;                       
          train with the loss $\mathcal{L}(\pmb{p}_i, \pmb{q}_i) + \lambda \mathcal{L}(\pmb{p}_i, \pmb{y}_i)$
}
\end{tabular}
\vspace{-0.2cm}
\caption{Introspective Learning with Online Self-explanations}
\label{alg}
\end{algorithm}

In the first training stage, the network is trained with one-hot labels. The goal is to initialize the network to a good starting point for generating meaningful explanations. Without warming up, the created explanations are almost random initially, which will mislead the training process.

In the second training stage, the network is trained with online soft labels extracted from online self-explanations. Given a training sample $(\pmb{x}_i, \pmb{y}_i)$, without loss of generality, we assume the $c$-th class is the ground truth $y^c_i=1$. In each epoch, we first classify the sample $\pmb{p}_i = f(\pmb{x}_i)$ and create explanations $\pmb{E}_i$ for every classification prediction. We then create new targets $\pmb{q}_i = (q_i^1, q_i^2, \cdots, q_i^c)$ to compute the loss instead of using the one-hot labels or soft labels provided by a teacher. The new targets $\pmb{q}_i$ are based on the explanations $\pmb{E}_i$ and the one-hot labels $\pmb{y}_i$. The one corresponds to the ground-truth class is specified with a pre-defined value $q_i^c=\alpha$, and the ones correspond to other classes are computed as
\begin{equation}
q_i^k = (1-\alpha)* \frac{cos(\pmb{e}_i^k, \pmb{e}_i^c)+1}{\sum_{m=1}^{c-1} (cos(\pmb{e}_i^m, \pmb{e}_i^c) +1)}
\end{equation}
where $k \neq c$. Since $cos(\pmb{e}_i^k, \pmb{e}_i^c) \in [-1, 1]$, 1 is added to the similarity value for numerical stability. The training process is also regularized by a cross-entropy term with one-hot labels and a common weight-decay term.

\textbf{Rationale behind the proposed learning procedure:} Our proposed learning procedure is motivated by the conclusions drawn in Sec. \ref{sec:dark}. We propose new targets to guide the training of the network. With the proposed targets, the responses in non-ground-truth classes (wrong responses) prevent the network from becoming over-confident. The wrong responses are not randomly computed but with the information of explanations. As the similarity between classes corresponds to the similarity between their explanations, the proposed targets contain class-similarity information. Therefore, our learning procedure includes effective components of KD without requiring a powerful teacher model. In other words, our proposed learning procedure can be seen as distilling knowledge directly from online explanations.

\textbf{Computational cost of the proposed learning procedure:} In our proposed algorithm, the extra cost to compute the explanations is required. For each training sample in a mini-batch, $c$ explanations are required to compute the corresponding targets. Using the implementation trick in \cite{goodfellow2015efficient}, all $c$ explanations can be obtained in a single backward pass. Hence, compared to the computationally expensive teacher model in KD, the extra cost brought by our learning procedure is much less.

\section{Experiments}
\label{sec:exp}

Following the previous work \cite{mirzadeh2019improved}, we use ResNet \cite{He2015DeepRL} in our experiment. On CIFAR10, the standard data augmentation is applied as in Sec.~\ref{sec:dark}. The models are trained with a batch size of 128 for 160 epochs using SGD. We start with a learning rate of 0.1, divide it by 10 at the $80$-th and the $120$-th epoch. In the proposed algorithm, we set the smoothing factor $\alpha = 0.9$ and the regularization strength $\lambda=0.1$.

The method we use to create explanations is the commonly used \textit{Grad-CAM}. It requires to specify a layer to create explanations. And we choose the last convolutional layer. The effectiveness of different explanation methods is also analyzed in Sec.~\ref{sec:xai}.

We compare the model trained with our proposed algorithm with the ones trained with different regularizations, the ones learned from peer networks, and the ones learned from a strong teacher. We further investigate our proposal with ablation studies in order to validate our hypothesis in the paper.

\subsection{Networks Trained with Regularizations}
The proposed algorithm can be viewed as a regularization method. It regularizes the networks so that their predictions are consistent with their explanations. In this experiment, we compare our method with other regularization methods, including Dropout, Label Smoothing (LS), and Confidence Penalty (CP). 

For Dropout, we drop neurons of the last layer of each block-layer with the drop rate in the range [0,1, 0.25, 0.5, 0.75], where 0.1 corresponds to the best performance. For LS, we vary the smoothing factor over [0.6, 0.7, 0.8, 0.9], where a factor of 0.9 turns out to be the best. For CP, we vary the penalty strength over [0.1, 0.3, 0.5, 1.0, 2.0, 4.0, 8.0], where the value of 0.5 gives the best result. The corresponding performance is listed in Table \ref{tab:reg}.

\begin{table*}[t]
        \centering
          \caption{Test accuracy (\%) of CNNs trained with different regularizations on CIFAR10 dataset. Our proposed LE outperforms the standard training procedure CE and others with various regularization methods.}
        \begin{tabular}{p{1.8cm} p{1.8cm}  p{1.8cm}  p{1.8cm}  p{1.8cm} p{1.8cm}  p{1.8cm}}
         \specialrule{.1em}{.05em}{.1em}
        Model & CE & Dropout & LS & CP & LE{\tiny (\textit{Grad-CAM})} \\
          \hline
        ResNet14 &  90,26{\tiny($\pm$0.33)} &   90,33{\tiny($\pm$0.15)} &  90,52{\tiny($\pm$0.12)}	&  90,42{\tiny($\pm$0.14)}	 &   \textbf{91,21}{\tiny($\pm$0.07)} \\
          \hline
        ResNet8   & 87,95{\tiny($\pm$0.32)} &  88,02{\tiny($\pm$0.02)}  &	88,14{\tiny($\pm$0.16)}   &  88,25{\tiny($\pm$0.13)} & \textbf{88,70}{\tiny($\pm$0.18)} \\
          \specialrule{.1em}{.05em}{.1em}
  \end{tabular}   
\label{tab:reg}
\end{table*}

All the regularizations outperform the baseline CE by alleviating the overfitting problem. Meanwhile, our method LE is superior to other regularization methods. LS is similar to ours, which is expected, as it also has responses in non-ground-truth classes in its specified targets. However, in LS, the specified response for each non-ground-truth class is the same. It results in the loss of information in the logits about the similarity between classes \cite{Muller2019WhenDL}. Different from LS, the specified targets in our method LE include the class-similarity information.

Following the research line, we also compare our method with others in the setting of previous work \cite{pereyra2017regularizing,alemi2017deep}. Concretely speaking, an Multilayer Perceptron (MLP) with fully connected layers of the form 784-1024-1024-10 and ReLU activations is trained on the MNIST dataset \cite{lecun1998gradient}. All models are trained with a batch size of 16 for 50 epochs using SGD with a learning rate of 0.01 and moment 0.9.

When training MLPs with our proposed algorithm, we applied \textit{Grad} approach to create explanations since \textit{Grad-CAM} is only specific for CNNs. 10 epochs are used for warm-up training in our learning procedure. For other regularizations, we explore their hyper-parameters in the same space as above and show the best of the exploration.

\begin{table*}[t]
        \centering
          \caption{Test error (\%) of MLPs trained with various regularizations on MNIST dataset. In our implementation, our LE achieves the best with tiny variance.}
        \begin{tabular}{c p{1.7cm}  p{1.7cm}  p{1.7cm}  p{1.7cm}  p{1.7cm}  p{1.7cm}}
         \specialrule{.1em}{.05em}{.1em}
        Implementations & CE & Dropout & LS & CP  & LE{\tiny (\textit{Grad})} \\
          \hline
       \cite{pereyra2017regularizing} & - & 1.28{\tiny($\pm$0.06)} & 1.23{\tiny($\pm$0.06)}	&  1.17{\tiny($\pm$0.06)} &  - \\
          \hline
        \cite{alemi2017deep}   & 1.38 &  1.34  &	1.40 &	1.36   & - \\
         \hline
        Ours   & 1.39{\tiny ($\pm$ 0.05)} &  1.27{\tiny($\pm$0.02)}  &	1.25{\tiny($\pm$0.02)} &	1.18{\tiny($\pm$0.06)}   & \textbf{1,10}{\tiny($\pm$0.03)} \\
          \specialrule{.1em}{.05em}{.1em}
  \end{tabular}   
\label{tab:reg_mlp}
\end{table*}

The comparison results in this simple setting are shown in Table \ref{tab:reg_mlp}. The scores of the first two rows are taken from their papers. The third row shows our implementations. In this simple setting with MLP, our method LE also outperforms other regularization methods. We only compare with the popular and state-of-the-art regularization methods and the ones related to our work. 
The score of VIB regularization proposed in \cite{alemi2017deep} is not listed since the setting is slightly different. Even though they require 12,03\% more parameters and 200 epochs to train (test error 1,13\%), our score still slight outperform theirs.

\begin{figure}[!htb]
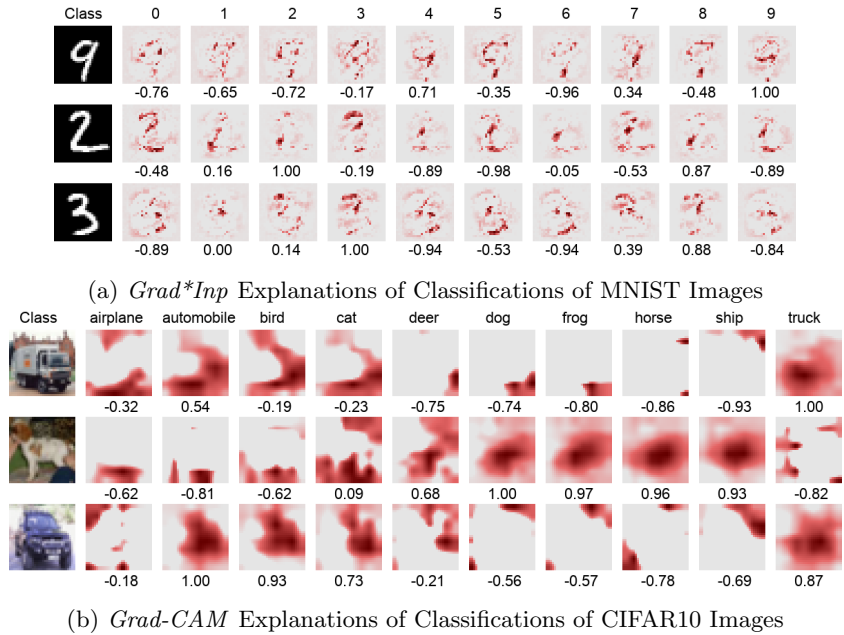

     \centering
     \begin{subfigure}[b]{\textwidth}
         \centering
         \includegraphics[scale=0.26]{exp_mnist}
         \caption{\textit{Grad*Inp} Explanations of Classifications of MNIST Images}
         \label{fig:exp_mnist}
     \end{subfigure}
     \begin{subfigure}[b]{\textwidth}
         \centering
         \includegraphics[scale=0.26]{exp_cifar}
         \caption{\textit{Grad-CAM} Explanations of Classifications of CIFAR10 Images}
         \label{fig:exp_cifar}
     \end{subfigure}
        \caption{The explanations and the similarity scores: For each image, we generate an explanation for each output class. The score under each explanation is the similarity score between the corresponding class and the ground truth class.}
        \label{fig:exp_vis}
\end{figure}

We also visualize the created explanations during training in Figure \ref{fig:exp_vis}. In each row, the first column shows the original image, and others columns show explanations corresponding to different classes $\pmb{E}_i = (\pmb{e}_i^1, \pmb{e}_i^2, \cdots, \pmb{e}_i^{10})$ (10 classes both in MNIST and in CIFAR10).

From Figure~\ref{fig:exp_vis}, we find the generated explanations is consistent with human's prior knowledge. 
For instance, in the first row of Figure~\ref{fig:exp_mnist}, the score under the explanation of the last column (the ground-truth class) is $cos(\pmb{e}_i^*, \pmb{e}_i^*)=1$. In the explanations corresponding to digits of $4$ and $7$, part of input features supports the corresponding scores. 
the similarity between the explanation of digit $4$ and the explanation of the digit $9$ (ground-truth class) is high. Besides, the similarity between the explanation of digit $7$ and the explanation of the digit $9$ is also high. 
Also, In the first row of Figure \ref{fig:exp_cifar}, the explanation of the ground-truth class focuses on the pixels of the truck, while others focus on irrelevant pixels. The explanation of $automobile$ class is more similar to that of the ground-truth class, and the corresponding cosine distance score is higher than that of others.

\subsection{Networks Learned from Peer Networks}

In our proposed algorithm, we trains a network with knowledge in explanations created on the same network. In this section, we compare the model trained with our learning procedure with the ones learned from peer networks.

We first train a ResNet8 using one-hot labels. Using its outputs as targets, we then train another ResNet8 from scratch. The second network is called Born-Again Network (BAN). We explore the temperature in the range of [1, 2, 5, 10, 20]. The training process can also be regularized by a cross-entropy term computed with one-hot true labels (BAN+L). The regularization strength is $\lambda = 0.1$. Similarly, we conduct experiments on ResNet14.

Another way to learn from peer networks is to train two networks at the same time. The loss of each network consists of two terms: a cross-entropy term computed with one-hot labels and a regularization term corresponding to the KL distance between outputs of the two networks. The regularization strength is $\lambda = 0.1$. With such loss, the two networks learn from each other, which is called Deep Mutual Learning (DML).

\begin{table*}[!htb]
        \centering
          \caption{Test accuracy (\%) of CNNs learned from peer networks on CIFAR10. Compared to those stuents trained with a peer network (i.e., a network with the same structure), our learning procedure LE{\tiny (\textit{Grad-CAM})} shows the best.}
        \begin{tabular}{l p{1.8cm} p{1.8cm} p{1.8cm} p{1.8cm} p{1.8cm}}
         \specialrule{.1em}{.05em}{.1em}
        Model      & CE  &  DML   & BAN   & BAN+L  & LE{\tiny (\textit{Grad-CAM})}    \\
          \hline
        ResNet14 &  90,26{\tiny($\pm$0.33)}  & 90,39{\tiny($\pm$0.15)} & 90,58{\tiny($\pm$0.15)} & 90,83{\tiny($\pm$0.28)}  & \textbf{91,21}{\tiny($\pm$0.07)} \\
          \hline
        ResNet8  & 87,95{\tiny($\pm$0.32)}    & 88,23{\tiny($\pm$0.27)} & 87,22{\tiny($\pm$0.03)} & 88,03{\tiny($\pm$0.33)} & \textbf{88,70}{\tiny($\pm$0.18)} \\
          \specialrule{.1em}{.05em}{.1em}
  \end{tabular}   
\label{tab:peer}
\end{table*}

The test accuracy is shown in Table \ref{tab:peer}. In BAN, although the two ResNet8 networks have the same representation capacity, it outperforms the CE. Both training procedures (BAN and DML) show better accuracy than the baseline CE without requiring a teacher. However, a peer network is still required. In contrast, our learning procedure shows better performance without a teacher and a peer.

\subsection{Networks Learned from Teacher Networks}
A large number of publications focus on the dark knowledge to be transferred from a teacher to a student, where Hinton et al. first represent the knowledge with soft labels (outputs of the teacher) as KD \cite{hinton2015distilling}. In orther works, the knowledge is presented by the intermediate representations (Hint) \cite{Romero2014FitNetsHF}, teacher's attention(AT) \cite{Zagoruyko2016PayingMA} or Flow of solution procedure (FSP) \cite{yim2017gift}. 

The setting in each method is as follows: Hint) we take the representation of the 2nd layer of ResNet26 as the Hint. AT) We use activation-based attentions where activation of feature maps are averaged over the channel dimension and normalized. The attention in all three layers of ResNet26 is captured to guide the training of the student. FSP) The flows between the 1st and the 2nd layers and between the 2nd and the 3rd layers are transferred from the teacher ResNet26 to the student. 
Following the experimental setup in these papers, we combine AT, FSP with Hinton's KD to achieve better distillation effectiveness.

Another way to learn from the teacher is to learn with the help of an assistant (TAKD) \cite{mirzadeh2019improved}. The learning procedure first distills the knowledge from the teacher to an assistant and then distill knowledge from the assistant to a student. In our experiments, we apply Hinton's distillation method in both distillation stages. I.e., the teachers, the assistants and the students are (ResNet26 $\rightarrow$ ResNet20 $\rightarrow$ ResNet14) and (ResNet26 $\rightarrow$ ResNet14 $\rightarrow$ ResNet8), respectively.

In all distillation methods, similar to \cite{crowley2018moonshine}, the temperature is set to 4, and $\lambda=0.1$ for the regularization corresponding to a cross-entropy term with one-hot labels. All other settings follow the original papers.

\begin{table*}[!htb]
        \centering
          \caption{Test accuracy (\%) of CNNs learned from more powerful teacher networks on CIFAR10: The teacher ResNet26 achieves 91,51\%. Compared to those students trained with a teacher, Our learning procedure LE{\tiny (\textit{Grad-CAM})} achieves competitive performance requiring no teacher.}
        \begin{tabular}{ l c c c c c c}
         \specialrule{.1em}{.05em}{.1em}
        Model & KD & Hint & AT\_Hinton & FSP\_Hinton  & TAKD & LE{\tiny (\textit{Grad-CAM})} \\
          \hline
          ResNet14 &  91,14{\tiny($\pm$0.19)} & 91,29{\tiny($\pm$0.19)} &  91,31{\tiny($\pm$0.19)} & 	\textbf{91,42}{\tiny($\pm$0.19)} &  91,27{\tiny($\pm$0.19)} &  91,21 {\tiny  ($\pm$ 0.07)} \\
          \hline
         ResNet8   & 88,07{\tiny($\pm$0.36)} & 88,52{\tiny($\pm$0.19)} & 88,15{\tiny($\pm$0.19)} &	88,11{\tiny($\pm$0.19)} & 88,09{\tiny($\pm$0.19)}  & \textbf{88,70}{\tiny($\pm$0.18)}   \\
          \specialrule{.1em}{.05em}{.1em}
  \end{tabular}   
\label{tab:teacher}
\end{table*}

The test accuracy of each distillation method is shown in Table \ref{tab:teacher}. Our proposed algorithm outperforms most of the distillation methods. Furthermore, without requiring a separate teacher network, LE is also comparable to the best method. We found that the rank of different KD methods is sensitive to the experimental setting. Our learning procedure achieves competitive performance without the cost to search for a suitable teacher model.

\subsection{Ablation Study}
\label{sec:xai}

\subsubsection{Class-similarity Information in Explanations}
We hypothesis that class-similarity information in explanations contributes to the success of the proposed algorithm is validated through the ablation study. In the second training stage of Algorithm \ref{alg}, we create new targets $\pmb{q}_i = (q_i^1, q_i^2, \cdots, q_i^c)$ using explanations. The responses in non-ground-truth classes contain class-similarity information. In the experiments, we permute the elements in $\pmb{q}_i$ except for the one corresponding to ground-truth classes (LE-Permut). By doing this, the class-similarity information provided by explanations is removed. The performance is shown in Table \ref{tab:xai_abl}.

\begin{table}[!htb]
        \centering
          \caption{Test accuracy (\%) of models trained with the proposed learning procedure on MNIST and CIFAR10 datasets.}
        \begin{tabular}{c     |  p{1.8cm}     |  p{1.8cm}       p{1.8cm}}
       \specialrule{.1em}{.05em}{.1em} 
         Datasets    & MNIST & \multicolumn{2}{c}{CIFAR10} \\
         \hline
         Models    & MLP1024 & \multicolumn{1}{c|}{ResNet14} & ResNet8 \\
       \specialrule{.1em}{.05em}{.05em} 
         CE                 & 98,61{\tiny($\pm$0.05)} & \multicolumn{1}{c|}{90,26{\tiny($\pm$0.33)}} & 87,95{\tiny($\pm$0.32)} \\
         \hline
         LE-Permut    & 98,81{\tiny($\pm$0.04)} & \multicolumn{1}{c|}{90,91{\tiny($\pm$0.16)}} & 88,22{\tiny($\pm$0.05)} \\
         \hline
         LE(\textit{Grad})     & 98,90{\tiny($\pm$0.03)} & \multicolumn{1}{c|}{91,01{\tiny($\pm$0.12)}} & 88,27{\tiny($\pm$0.05)} \\
         \hline
         LE(\textit{Grad-CAM})  & - & \multicolumn{1}{c|}{91.21{\tiny($\pm$0.07)}} & 88,70{\tiny($\pm$0.18)} \\
        \specialrule{.1em}{.05em}{.1em}
  \end{tabular}   
\label{tab:xai_abl}
\end{table}

Compared with LE-Permut, both LE(\textit{Grad}) and LE(\textit{Grad-CAM}) show better performance. The observation indicates that class-similarity information is an essential component of the proposed algorithm. Another effective component that contributes to the our algorithm's effectiveness is the responses in the non-ground-truth classes, which is shown by the observation, that LE-Permut outperforms the baseline CE without similarity information. 

Interestingly, LE-Permut also outperforms the Label Smoothing (LS) (90,52{\tiny ($\pm$ 0.12)}\% on ResNet14, 88,14{\tiny ($\pm$ 0.16)}\% on ResNet8). LS sets the same response for all the non-ground-truth classes, which leads to the fact that the learned representations of a class have the same distance as those of other classes. The responses in non-ground-truth classes in LS prevents overfitting problem. However, the symmetry property constrains the learning of representations. By randomly specify the response in the non-ground-truth classes, LE-Permut prevents both overfitting problem and the symmetry problem.

\begin{figure*}[!htb]
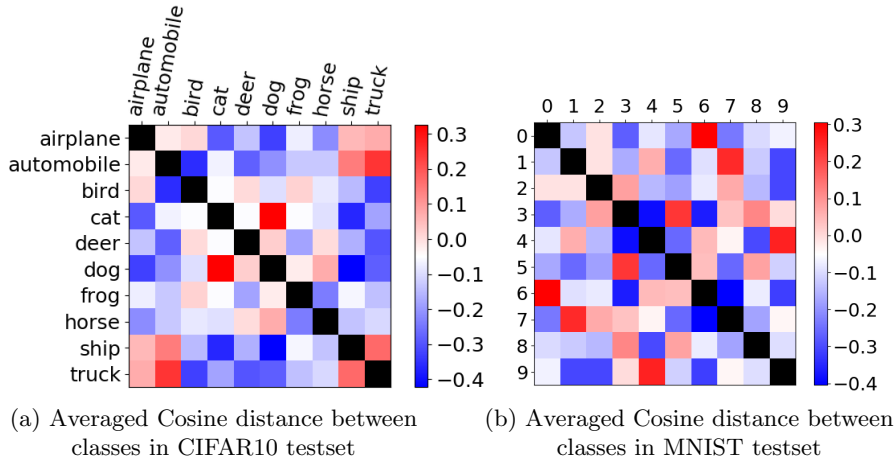

     \centering
     \begin{subfigure}[b]{0.48\textwidth}
         \centering
         \includegraphics[scale=0.42]{sim_cifar}
         \caption{Averaged Cosine distance between classes in CIFAR10 testset}
         \label{fig:matrix_cifar}
     \end{subfigure}
      \hspace{0.25cm}
     \begin{subfigure}[b]{0.48\textwidth}
         \centering
         \includegraphics[scale=0.32]{sim_mnist}
         \caption{Averaged Cosine distance between classes in MNIST testset}
         \label{fig:matrix_mnist}
     \end{subfigure}
        \caption{Visualization of class-similarity information in explanations}
        \label{fig:matrix_vis}
\end{figure*}

We also visualize the class-similarity information in Figure \ref{fig:matrix_vis}. Each grid corresponds to the cosine distance between two explanations of two classes. The distance scores are averaged across the images in the test dataset. In Figure \ref{fig:matrix_cifar}, the two classes in each pair \{($truck$, $automobile$),  ($dog$, $cat$)\} are similar to each other. In Figure \ref{fig:matrix_mnist}, we observe that the two classes in each class pair $\{(0, 6), (1, 7), (3, 5), (4, 9)\}$ are more similar to each other than other class pairs. The observations are consistent with the human's prior knowledge. It is therefore not surprising that the model can be improved by such knowledge. 

\subsubsection{Learning with different Explanation Methods}
In most experiments above, we apply \textit{Grad-CAM} (one of the state-of-the-art methods) to create explanations. The menthod is shown to create class-discriminative explanations, which can better describe the relationship between classes than others. In this Section, we show how the effectiveness of our proposed learning procedure is affected by different saliency (explanation) methods.

In Section \ref{sec:relate}, we introduce four popular and efficient saliency methods to create explanations, \textit{Grad}, \textit{Grad*Input}, \textit{GuidedBP} and \textit{Grad-CAM}. The work \cite{adebayo2018sanity} using Spearman Rank Correlation to evaluates the class-discriminativeness of each saliency method quantitatively. The evaluation results shows that the methods can be ordered by the class-discriminativeness as: \textit{Grad*Input}$<$ \textit{GuidedBP} $<$ \textit{Grad} $<$ \textit{Grad-CAM}. 

\begin{figure*}[!htb]
     \centering
     \begin{subfigure}[b]{\textwidth}
         \centering
         \includegraphics[scale=0.5]{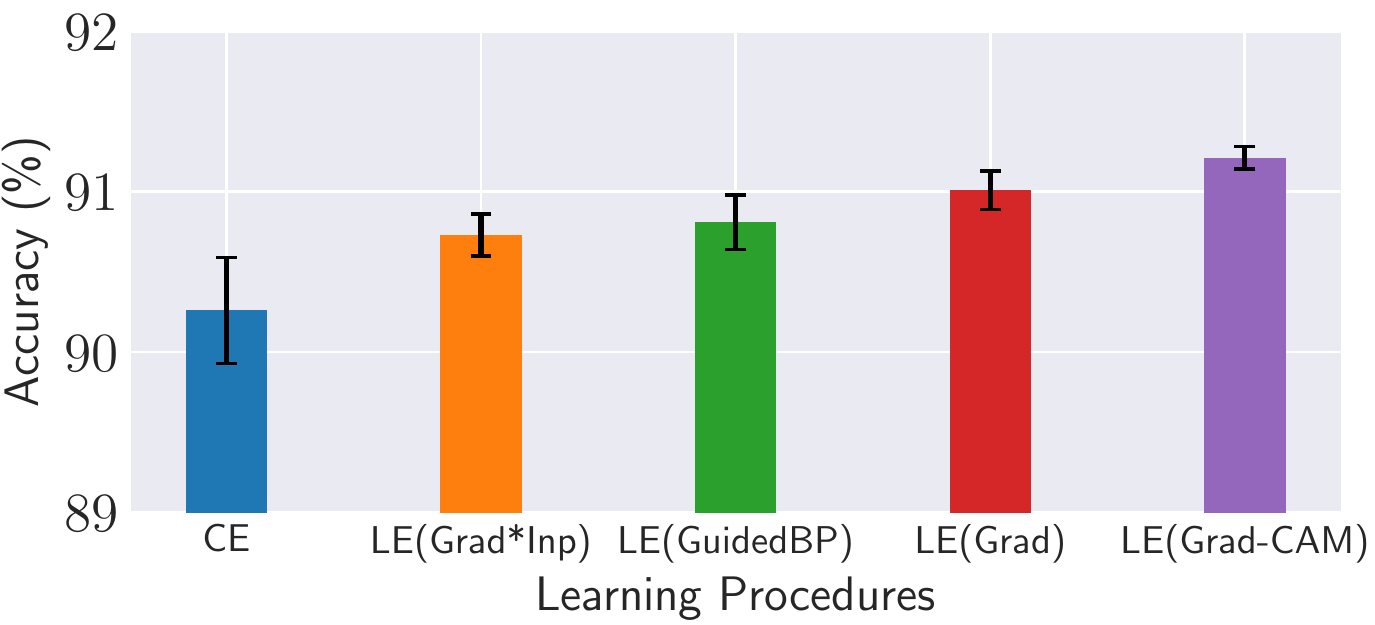}
         \caption{ResNet14 on CIFAR10 testset}
         \label{fig:matrix_cifar}
     \end{subfigure}
      \hspace{0.25cm}
      
     \begin{subfigure}[b]{\textwidth}
         \centering
         \includegraphics[scale=0.5]{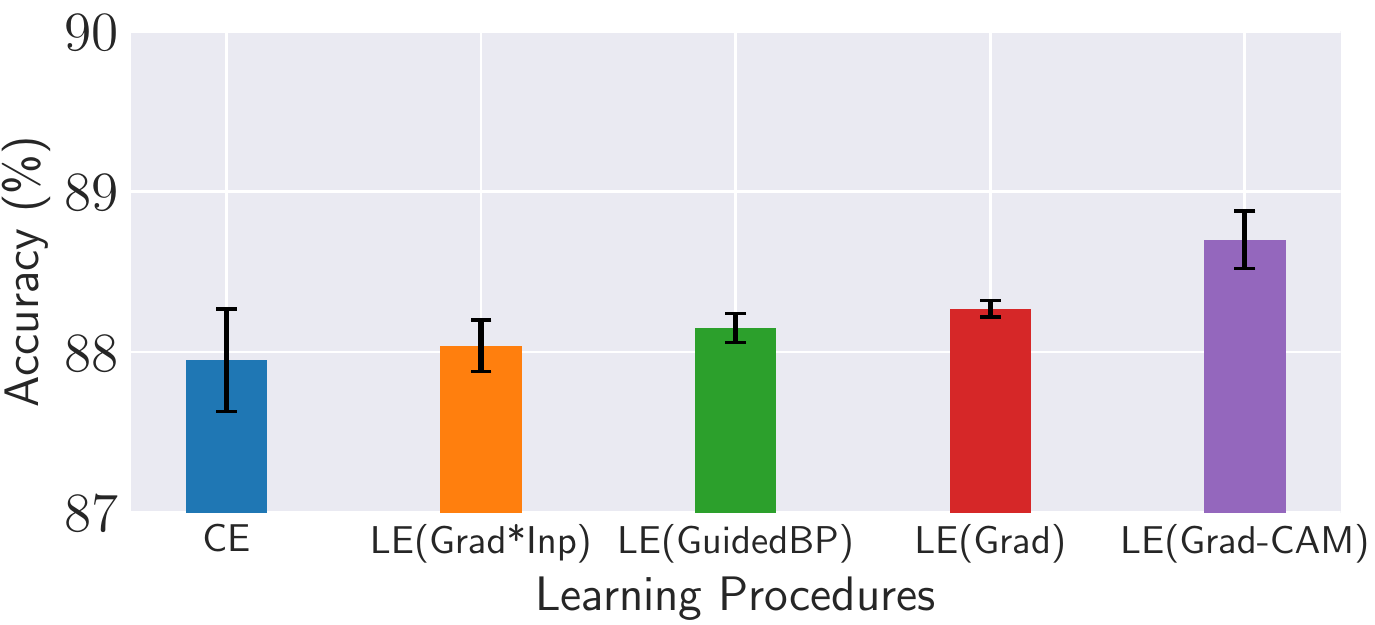}
         \caption{ResNet8 on CIFAR10 testset}
         \label{fig:matrix_mnist}
     \end{subfigure}
        \caption{Test accuracy (\%) of models learned with the proposed learning procedure with different explanation methods: The one trained with the state-of-the-art explanation method (\textit{Grad-CAM}) shows the best performance.}
        \label{fig:xai_methods}
\end{figure*}

We apply them in the proposed learning procedure, respectively. The test accuracy of models is shown in Figure \ref{fig:xai_methods}. The order by class-discriminativeness is consistent with the order by corresponding test accuracy. When the saliency method is more class-discriminative, the class-similarity information in their explanations is more accurate, the performance of the corresponding model is better.

We claim that the similarity information between classes is an effective component of KD. Our learning procedure integrates the effective component by extracting similarity information from self-explanations. This experiment further confirms our claims.

\section{Conclusion and Future Work}
Knowledge distillation (KD) explores the capacity of neural networks with privileged information provided by a strong teacher network. This work reveals the effective components of KD. Motivated by the findings, we propose an introspective learning algorithm to explore the network capacity by leveraging online self-explanation. The models trained with our proposed algorithm outperform those trained with the standard training procedure and various regularizations. Compared to the ones learned from peer networks or powerful teacher networks, our algorithm still shows competitive performance without peers or teachers.

To the best of our knowledge, this is the first work to leverage online explanations to improve the model training process. Our proposed algorithm illustrates one way to extract knowledge from saliency map explanations. The explanations beyond saliency maps \cite{gu2019semantics,kim2018interpretability,gu2019neural} can also be applied to implement introspective learning in a different way. We leave other possibilities to learn with explanations in future investigation.

%===========================================================
\bibliographystyle{splncs}
\bibliography{egbib}

\end{document}